\documentclass[conference]{IEEEtran}
\IEEEoverridecommandlockouts
\usepackage{cite}
\usepackage{amsmath,amssymb,amsfonts}
\usepackage[cmintegrals]{newtxmath}
\usepackage{algorithmic}
\usepackage{graphicx}
\usepackage{float}
\usepackage{caption}
\usepackage{subcaption}
\usepackage{textcomp}
\usepackage{xcolor}
\usepackage{multirow}
\usepackage{booktabs}
\usepackage[ruled, vlined]{algorithm2e}

\allowdisplaybreaks
\def\BibTeX{{\rm B\kern-.05em{\sc i\kern-.025em b}\kern-.08em
    T\kern-.1667em\lower.7ex\hbox{E}\kern-.125emX}}
\begin{document}

\title{Multi-Label Graph Convolutional Network Representation Learning}

\author{
\IEEEauthorblockN{Min Shi\textsuperscript{1}, Yufei Tang\textsuperscript{1}, Xingquan Zhu\textsuperscript{1} and Jianxun Liu\textsuperscript{2}}
\IEEEauthorblockA{
\textsuperscript{1} Department of Computer \& Electrical Engineering and Computer Science, Florida Atlantic University, USA \\
\textsuperscript{2} School of Computer Science and Engineering, Hunan University of Science and Technology, China
\\ Email: \{mshi2018, tangy, xzhu3\}@fau.edu, ljx529@gmail.com}
}

\maketitle

\begin{abstract}
Knowledge representation of graph-based systems is fundamental across many disciplines. To date, most existing methods for representation learning primarily focus on networks with simplex labels, yet real-world objects (nodes) are inherently complex in nature and often contain rich semantics or labels, $e.g.$, a user may belong to diverse interest groups of a social network, resulting in multi-label networks for many applications. The multi-label network nodes not only have multiple labels for each node, such labels are often highly correlated making existing methods ineffective or fail to handle such correlation for node representation learning. In this paper, we propose a novel multi-label graph convolutional network (ML-GCN) for learning node representation for multi-label networks. To fully explore label-label correlation and network topology structures, we propose to model a multi-label network as two Siamese GCNs: a node-node-label graph and a label-label-node graph. The two GCNs each handle one aspect of representation learning for nodes and labels, respectively, and they are seamlessly integrated under one objective function. The learned label representations can effectively preserve the inner-label interaction and node label properties, and are then aggregated to enhance the node representation learning under a unified training framework. Experiments and comparisons on multi-label node classification validate the effectiveness of our proposed approach.

\end{abstract}


\section{Introduction}
Graphs have become increasingly common structures for organizing data in many complex systems such as sensor networks, citation networks, social networks and many more \cite{zhang2018network}. Such a development raised new requirement of efficient network representation or embedding learning algorithms for various real-world applications, which seeks to learn low-dimensional vector representations of all nodes with preserved graph topology structures, such as edge links, degrees, and communities \textit{etc}. The graph edges inherently reflect semantic relevance between nodes, where nodes with similar neighborhood structures usually tend to share identical labeling information, \textit{i.e.}, clustering together characterized by a single grouping label. For examples, in a scientific collaboration network, two connected authors often belong to a common area of science \cite{2,3}, and in a protein-protein interaction network, proteins co-appeared in many identical  protein complexes are likely to serve with the similar biological functions.

To date, a large body of work has been focused on the representation learning of graphs with simplex labels \cite{4,5}, where each node only has a single label which is used to model node relationships, \textit{i.e.}, two nodes in a neighborhood are forced to have an identical unique label in the learning process. However, graph nodes associated with multiple labels are ubiquitous in many real-world applications. For example, in an image network, a photograph can belong to more than one semantic class, such as \textit{sunsets} and \textit{beaches}. In a patient social network, a patient may be suffering from \textit{diabetes} and \textit{kidney cancer} at the same time. Similarly, in many social networks, such as BlogCatalog and Flickr, users are allowed to join various groups that respectively represent their multiple interests. For all these networks, each node not only has content (or features), it is also associated with multiple class labels. 

In general, multi-label graphs primarily differ from simplex-label graphs in twofold. First, every node in a multi-label graph could be associated with a set of labels, thus graph structures usually encode much more complicated relationships between nodes with shared labels, \textit{i.e.}, an edge could either reflect a simple relationship of some single label or interpret a very complex relationship of multiple combined labels. Second, it has been widely accepted that label correlations and dependencies are widespread between multiple labels \cite{6,7}, \textit{i.e.}, the \textit{sunsets} are frequently correlated with the \textit{beaches}, and \textit{diabetes} could finally lead to \textit{kidney cancer}. Therefore, the correlation and interaction between labels could provide implicit and supplemental factors to enhance and differentiate node relationships that cannot be explicitly captured by the discrete and independent labels in a simplex-label graph.

\begin{figure}
  \centering
    \includegraphics[width=0.44\textwidth]{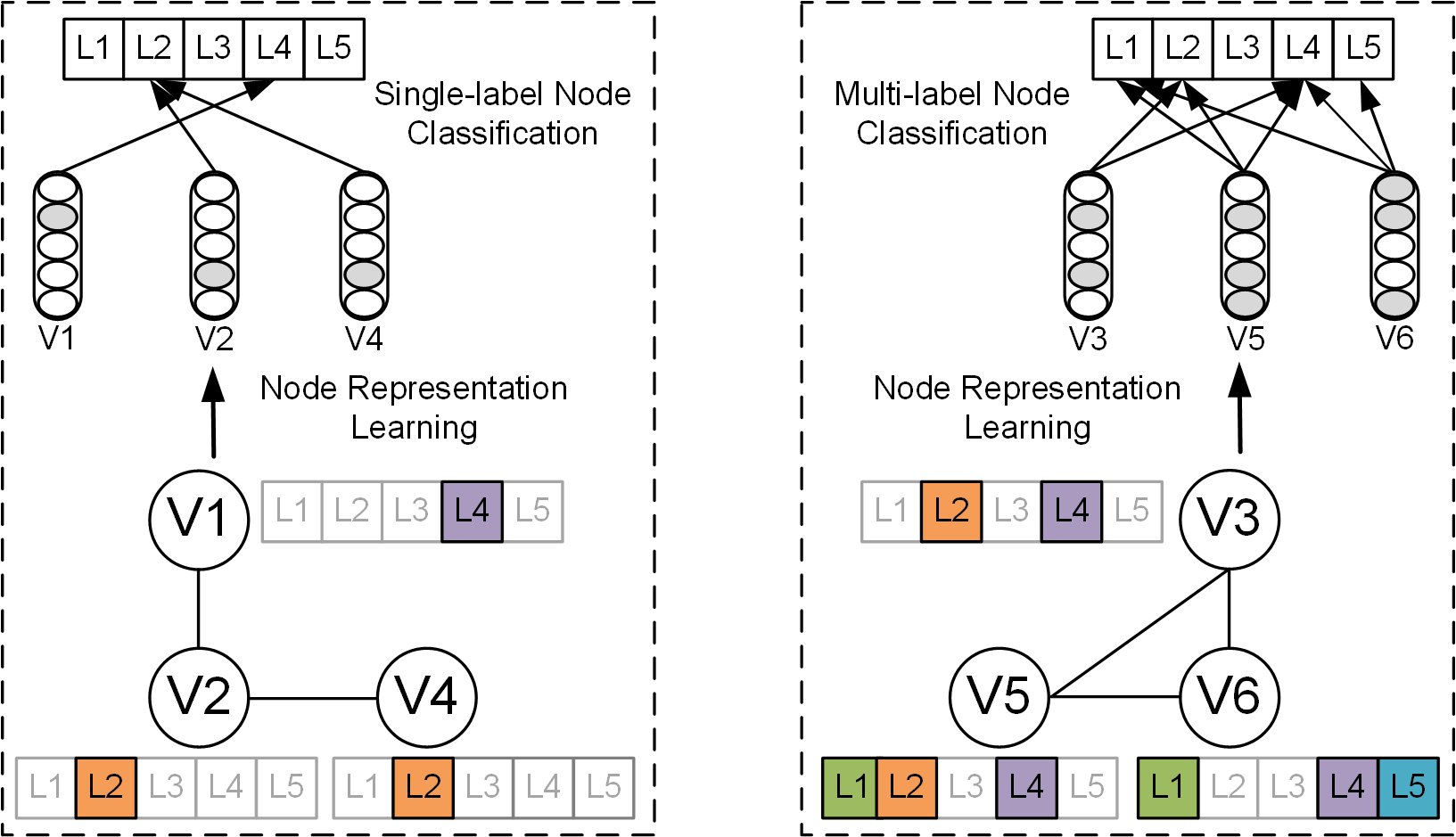}
  \caption{Illustration of the difference between simplex-label graph learning \textit{vs.} multi-label graph learning, where the labels of each node are highlighted. In a simplex-label graph, each node is associated with only one label and performs single-label classification based on learned representations. In a multi-label graph, each node may be associated with multiple labels and these node labels are often highly correlated to represent node semantics.}
  \label{fig:motivation}
\end{figure}

Indeed, multi-label learning is a fundamental problem in the machine learning community \cite{8}, with significant attentions in many research domains such as computer vision \cite{9}, text classification \cite{10} and tag recommendation \cite{17}. However, research on multi-label graph learning is still in its infancy. Existing methods either consider graphs with simplex labels \cite{5,11} or treat multiple labels as plain attribute information to enhance the graph learning process \cite{12,13}. Such learning paradigms, however, neglect the fact that the information of one label may be helpful for the learning of another related label \cite{6}---the label correlations may provide helpful extra information especially when some labels have insufficient training examples. To address this constrain and meanwhile move forward the graph learning theory a litter, we propose multi-label graph representation learning in this paper, where each node has a collection of features as well as a set of labels. Figure~\ref{fig:motivation} illustrates the difference between our studied problem and the traditional simplex-label graph learning. We argue the key for multi-label graph learning is to efficiently combine network structures, node features and label correlations for enriched node relationships modeling in a mutually reinforced manner.

Incorporating labels and their correlations with graph structures for graph representation learning is a nontrivial task. First, in a multi-label graph, two linked nodes may share one or multiple identical labels, thus their affinity cannot be simply determined by one observed edge that is indistinguishable from others. Second, while each label can be seen as an abstraction of nodes sharing similar network structures and features, the label-label correlations would bring about dramatic impact on the node-node interactions, thus it is hard to constrain and balance the two aspects of relations modeling for an optimal graph representation learning as a whole. Recently, a general class of neural network called Graph Convolutional Networks (GCN) \cite{14} shows good performance for learning node representations from graph structures and features by performing the supervised single-label node classification training. GCN operates directly on a graph and induces embedding vectors of nodes based on the spectral convolutional filter that enforces each node to aggregate features from all neighbors to form its representation. 

In this paper, we advance this emerging tool to multi-label node classification and propose a novel model called Multi-Label GCN (ML-GCN) to specifically handle the multi-label graph learning problem. ML-GCN contains two Siamese GCNs to learn label and node representations from a high-layer label-label-node graph and a low-layer node-node-label graph, respectively. The high-layer graph learning serves to model label correlations, which only updates the label representations with preserved labels, label correlations and node community information by performing a single-label classification. The derived label representations are subsequently aggregated to enhance the low-layer graph learning, which carries out node representation learning from graph structures and features by performing a multi-label classification. Learning in these two layers can enhance each other in an alternative training manner to optimize a collective classification objective.

Our main contributions are summarized as follows: 

\begin{enumerate}
    \item We advance the traditional simplex-label graph learning to a multi-label graph learning setting, which is more general and common in many real-world graph-based systems.
    \item Instead of treating multiple labels as flat attributes, like many existing methods do, We propose to leverage label correlations to strengthen and differentiate edge relationships between nodes.
    \item We propose a novel model ML-GCN to handle multi-label graphs. It can simultaneously integrate graph structures, features, and label correlations for enhanced node representation learning and classification.
\end{enumerate}

The rest of this paper is organized as follows. Section II surveys the related work. Section III reviews some preliminaries, including definition of the multi-label graph learning problem and the graph convolutional networks used in our approach. The proposed model for multi-label graph embedding is introduced in Section IV. Section V reports experiments and comparisons to validate the proposed approach. Finally, Section VI concludes the paper.

\section{Related Works}
This section presents existing works related to our studied problem in this paper, including multi-label learning and graph representation learning.

\subsection{Multi-label Learning}
Multi-label learning is a classical research problem in the machine learning community with applications ranging from document classification and gene function prediction to automatic image annotation \cite{15}. In a multi-label learning task, each instance is associated with multiple labels represented by a sparse label vector. The objective is to learn a classifier that can automatically assign an instance with the most relevant subset of labels \cite{8}. Techniques for multi-label classification learning can be broadly divided into two categories \cite{16}: the problem transformation-based and the algorithm adaption-based. The former class of methods generally transforms the multi-label classification task into a series of binary classification problems \cite{6,18} while adaption-based methods try to generalize some popular learning algorithms to enable a multi-label learning setting \cite{19,20}.

Multi-label learning methods for graph-based data did not attract much attention in the past. DeepWalk \cite{21} was proposed to learn graph representations that are then used for training a multi-label classifier. However, DeepWalk only exploits graph structures, with valuable label and label  correlation information not preserved in learned node embeddings. Wang et al. \cite{4} and Huang et al. \cite{13} proposed to leverage labeling information along with graph structures for enriched representation learning. However, these methods either consider simplex-label graphs or treat multiple labels as plain attribute genes to support graph structure modeling. Such paradigms still neglect frequent label correlations and dependencies which are demonstrably helpful properties in multi-label learning problems \cite{9,10}.

\subsection{Graph Representation Learning}
Graph representation learning~\cite{zhang2018network, wu2019comprehensive} seeks to learn low-dimensional vector representations of a given network, such that various downstream analytic tasks like link prediction and node classification can be benefited. Traditional methods in this area are generally developed based on shallow neural models, such as DeepWalk\cite{21}, Node2vec\cite{26} and LINE\cite{24}. To preserve the node neighborhood relationships, they typically perform truncated random walk over the whole graph to generate a collection of fixed-length node sequences, where nodes within the same sequences are assumed to have semantic connections and will be mapped to be close in the learned embedding space. However, above methods only consider modeling the edge links to constrain node relations, which may be insufficient especially when the network structures are very sparse. To mitigate this issue, many methods \cite{3, yang2015network} are proposed to additionally embed the rich network contents or features associated such as the user profiles in a social network and the publication descriptions in a citation network. For example, TriDNR \cite{pan2016tri} was proposed to learn simultaneously from the network structures and textual contents, where structures and texts are mutually boosted to collectively constrain the similarities between learned node representations. In general, most real-word graphs are sparse in connectivity (\textit{e.g.}, each node only connects several others in the huge node space), while node contents or features can be leveraged to either enhance node relevance or repair the missing links in the original network structures \cite{le2014probabilistic}.

However, above representation learning methods belong to the class of shallow neural models, which may have limitations in learning complex relational patterns between graph nodes. Recently, there is growing interest in adapting deep neural networks to handle the non-Euclidean graph data \cite{11,22}. Several works seek to apply the concepts of convolutional neural networks to process arbitrary graph structures \cite{14,24}, with GCN \cite{14} achieving state-of-the-art representation learning and node classification performance on a number of benchmark graph datasets. Following this success, Yao el al. \cite{25} proposed a Text GCN for document embedding and text classification based on a constructed heterogeneous word-document graph. Graph Attention Networks (GAN) \cite{11} are another recently proposed end-to-end neural network structure similar to GCNs, which introduce attention mechanisms that assign larger weights to the more important nodes, walks, or models. Inspired by these deep neural models targeted at mostly the simplex-label graphs, we generalize GCN and propose a novel training framework, ML-GCN, to address the multi-label graph learning problem in this paper.

\begin{figure*}
  \centering
    \includegraphics[width=0.96\textwidth]{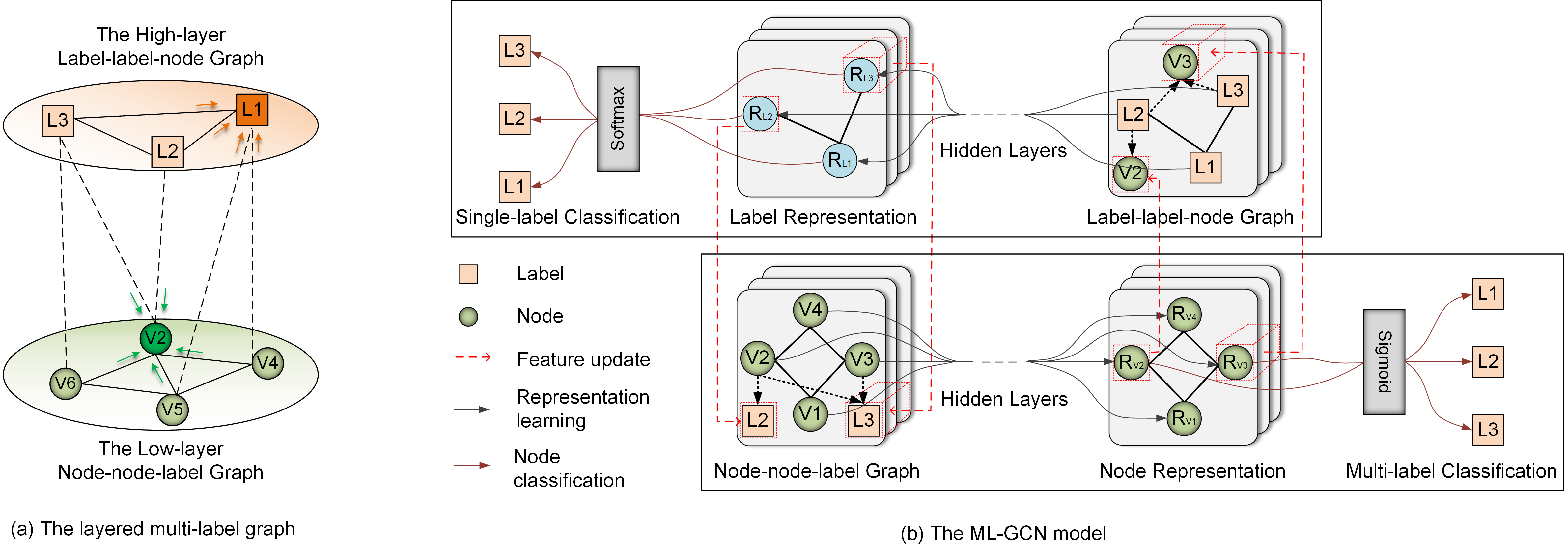}
  \caption{The proposed ML-GCN model for multi-label graph learning. (a) shows a multi-label graph organized in two layers. (b) shows the proposed architecture that contains two Siamese GCNs to learn from the label-label-node graph and node-node-label graph, respectively. The upper panel uses label-label-node graph to learn label representation (from right to left), and the lower panel uses node-node-label graph to learn node representation (from left to right).}
  \label{fig:mlgcn}
\end{figure*}

\section{Problem Definition \& Preliminaries}
\subsection{Problem Definition}
A multi-label graph is represented as $G=(\mathbf{v},\mathbf{e},\mathbf{l},\mathbf{A},\mathbf{B},\mathbf{X})$, where $\mathbf{v}=\{v_i\}_{i=1,\cdots,n}$ is a set of unique nodes, $\mathbf{e}=\{ e_{i,j} \}_{ {i,j}=1,\cdots,n;~i \neq j }$ is a set of edges and $\mathbf{l}=\{c_r\}_{r=1,\cdots,m}$ is a set of unique labels, respectively. $n$ is the total number of nodes in the graph and $m$ is the total number of labels in the labeling space. $\mathbf{A}$ is a $n \times n$ adjacency matrix with $\mathbf{A}_{i,j}=w_{i,j}>0$ if $e_{i,j}\in \mathbf{e}$ and $\mathbf{A}_{i,j}=0$ if $e_{i,j}\notin \mathbf{e}$. $\mathbf{B}$ is a $n \times m$ affiliation matrix of labels with $\mathbf{B}_{i,r}=1$ if $v_i$ has label $c_r \in \mathbf{l}$ or otherwise $\mathbf{B}_{i,r}=0$. Finally, $\mathbf{X} \in \mathbb{R}^{n \times d_i}$ is a matrix containing all $n$ nodes with their features, $i.e.$, $\mathbf{X}_i \in \mathbb{R}^{d_i}$ represents the feature vector of node $v_i$, where $d_i$ is the feature vector's dimension. 

In this paper, the multi-label graph learning \textbf{aims} to represent nodes of graph $G$ in a new $m$-dimensional feature space $\mathcal{H}^{m}$, embedding information of graph structures, features, labels and label correlations preserved, \textit{i.e.}, learning a mapping $f:G \rightarrow\{\mathbf{h}_i\}_{i=1,\cdots,n}$ such that $\mathbf{h}_i \in \mathcal{H}^{m}$ can be used to accurately infer labels associated with node $v_i$. 

\subsection{Graph Convolutional Networks}
GCN \cite{wu2019comprehensive} is a general class of convolutional neural networks that operate directly on graphs for node representation learning and classification by encoding both the graph structures and node features. In this paper, we focus on the spectral-based GCN \cite{15}, which assumes that neighborhood nodes tend to have identical labels guaranteed by each node gathers features from all neighbors to form its representation. Given a network $G=(\mathbf{v},\mathbf{e},\mathbf{X})$, which has $n$ nodes and each node has a set of $d_i$-dimensional features ($\mathbf{X}\in \mathbb{R}^{n \times {d_i}}$ denotes the feature vector matrix of all nodes), GCN takes this graph as input and obtains the new low-dimensional vector representations of all nodes though a convolutional learning process. More specifically, with one convolutional layer, GCN is able to preserve the 1-hop neighborhood relationships between nodes, where each node will be represented as a $m-$dimension vector. The output feature matrix for all nodes $\mathbf{X}^{(1)} \in \mathbb{R}^{n \times m}$ can be computed by:
\begin{equation}
\mathbf{X}^{(1)} = \rho (\tilde{\mathbf{A}} \mathbf{X}^{(0)} \mathbf{W}_{0})
\end{equation}
where $\tilde{\mathbf{A}}=\mathbf{D}^{-\frac{1}{2}}(\mathbf{I}+\mathbf{A})\mathbf{D}^{-\frac{1}{2}}$ is the normalized symmetric adjacency matrix. $\mathbf{D}$ is the degree matrix of $(\mathbf{I}+\mathbf{A})$ and $\mathbf{I}$ is an identity matrix with corresponding shape. $\mathbf{X}^{(0)} \in \mathbb{R}^{n \times d_i}$ is the input feature matrix (e.g., $\mathbf{X}^{(0)} = \mathbf{X}$ ) for GCN and $\mathbf{W}_{0} \in \mathbb{R}^{d_i\times m}$ is a weight matrix for the first convolutional layer. $\rho$ is an activation function such as the \textit{ReLU} represented by $\rho (x) = \max (0,x)$. If it is necessary to encode $k$-hop neighborhood relationships, one can easily stack multiple GCN layers, where the output node features of the $j$th (1 $\leq$ $j$ $\leq$ $k$) layer is calculated by:
\begin{equation}
\mathbf{X}^{(j+1)} = \rho (\tilde{\mathbf{A}} \mathbf{X}^{(j)} \mathbf{W}_{j})
\end{equation}
where $\mathbf{W}_j \in \mathbb{R}^{d_h \times d_h}$ is the weight matrix for the $j$th layer and $d_h$ is the feature vector dimension output in the hidden convolutional layer.

\section{The Proposed Approach}
In this section, we first present the proposed Multi-Label Graph Convolutional Networks (ML-GCN) model, where the node representations are learned and trained through the supervised classification. Then, we provide the training and optimization details which incorporate node and label representation learning by a collective objective, followed by the computation complexity analysis of the ML-GCN.
\subsection{Multi-Label Graph Convolutional Networks}
As discussed in previous sections, the key and challenge for multi-label graph learning are to simultaneously learn from the graph structures, features, labels and label correlations, where different aspects of learning could enhance each other to achieve a global good network representation. To support the incorporation of labels, we can simply build a heterogeneous node-label graph similar to the text GCN \cite{25}, where common nodes and label nodes are directly connected by their labeling relationships. However, such a diagram makes it hard to model the higher-order label correlations since labels must reach each other through common nodes, \textit{i.e.}, one cannot directly encode $k$-hop neighborhood node relations and label correlations by a GCN with $k$ convolutional layers. To enable immediate and flexible label interactions, we consider a stratified graph structure shown in Fig.~\ref{fig:mlgcn}(a), which is defined as follows:

\noindent \textbf{Label-label-node graph:} In this graph, labels connect each other by their co-occurrence relations, \textit{i.e.}, two labels have an edge if they appear at the same time in the label set of some common node. Meanwhile, common nodes are seen as attributes that link with label nodes based on their corresponding labeling relationships.

\noindent \textbf{Node-node-label graph:} In this graph, common nodes link together to form the original graph structure. Meanwhile, associated label nodes are seen as the attributes of the common node-node graph.

Such a construction of the layered multi-label graph in Fig.~\ref{fig:mlgcn}(a) could bring three main favorable properties. First, the label-label connectivity allows direct and efficient higher-order label interactions by simply adjusting the number of convolutional layers in GCN. In addition, common nodes as attributes of the label nodes enable to encode graph community information in learned label representations, as nodes with identical labels tend to form a cluster or community. Lastly, the learned node representations can naturally preserve labels, label correlations and graph community information by taking label nodes as attributes of the node-node graph.

Fig.~\ref{fig:mlgcn}(b) shows the proposed ML-GCN model that contains two Siamese GCNs to simultaneously learn the label and node representations from the given multi-label graph, where input feature vectors of both the attributed label and common nodes will be regularly updated during the training. First, the high-layer GCN learns label representations from the label-label-node graph through supervised single-label classification. Let $\mathbf{Y} \in \mathbb{R}^{m \times d_i}$ be the input feature matrix of all $m$ label nodes, $\mathbf{C}$ be the $m \times m$ adjacency matrix recording the co-occurrence relations between label nodes, and $\mathbf{F}$ be the $(n+m) \times (n+m)$ adjacency matrix of the input label-label-node graph. The first convolutional layer aggregates information from both the neighborhood label nodes and the associated common nodes (\textit{e.g.}, label node L1 in Fig.~\ref{fig:mlgcn}(a)), where the new $m$-dimensional label node feature matrix $\mathbf{L}^{(1)} \in \mathbb{R}^{n \times m}$ is computed by:
\begin{equation}
\mathbf{L}^{(1)} = \rho (\tilde{\mathbf{F}}^{*} \mathbf{Y}^{*} \mathbf{W}_{0}^{l})
\end{equation}
where $\rho$ is an activation function, such as the \textit{ReLU} represented by $\rho (x) = \max (0,x)$, $\mathbf{W}_{0}^{l} \in \mathbb{R}^{d_i\times m}$ is a weight matrix for the first label-label-node GCN layer and $\mathbf{Y}^{*}=[\mathbf{Y};\mathbf{X}]^{T}$ is a vertically stacked $(m+n)\times d_i$ feature matrix. $\tilde{\mathbf{F}}^{*}$ is a truncated normalized symmetric adjacency matrix obtained by:
\begin{equation}
\mathbf{F}^{*} = \mathbf{F}+\mathbf{I}_{m+n};  \tilde{\mathbf{F}} = \mathbf{D}_{f}^{-\frac{1}{2}} \mathbf{F}^{*} \mathbf{D}_{f}^{ - \frac{1}{2}}; \tilde{\mathbf{F}}^{*} = \tilde{\mathbf{F}}[:m]
\end{equation}
where $\mathbf{I}_{m+n}$ is the identity matrix, $\mathbf{D}_f$ is the degree matrix with $\mathbf{D}_{f,ii}=\sum_{j}\mathbf{F}^{*}_{ij}$. One layer GCN only incorporates immediate label node neighbors. When higher order label correlations need to be preserved, we can easily stack multiple GCN layers (\textit{e.g.}, the layer number $k \geq 2$) by:
\begin{equation}
\mathbf{L}^{(k)} = \rho (\tilde{\mathbf{C}}\mathbf{L}^{(k-1)} \mathbf{W}_k^l) 
\end{equation}
where $\tilde{\mathbf{C}} = \mathbf{D}_{c}^{-\frac{1}{2}} (\mathbf{C}+\mathbf{I}_m) \mathbf{D}_{c}^{ - \frac{1}{2}}$ is the normalized symmetric adjacency matrix and $\mathbf{D}_{c,ii}=\sum_{j}(\mathbf{C}+\mathbf{I}_m)_{ij}$. The last layer output label embeddings have the same size as the total number of labels, $m$, and are through a \textit{softmax} classifier to perform the single-label classification (\textit{e.g.}, assume we consider a two-layer GCN) by:
\begin{equation}
\mathbf{O}^l=\tilde{\mathbf{C}} ReLU (\tilde{\mathbf{F}}^* \mathbf{Y}^* \mathbf{W}_0^l)\mathbf{W}_1^l
\end{equation}
\begin{equation}
\mathbf{Z}^l=softmax(\mathbf{O}^l)=\frac{{\exp (\mathbf{o}^l )}}{{\sum\nolimits_i {\exp (\mathbf{o}_i^l )} }}
\end{equation}
where $\mathbf{W}_0^l \in \mathbb{R}^{d_i \times d_h}$ and $\mathbf{W}_1^l \in \mathbb{R}^{d_h \times m}$ are the weight matrices for the first and second label-label-node GCN layers, respectively. Let $\mathbf{Y}^l$ be the one-hot label indicator matrix of all label nodes, the classification loss can be defined as the cross-entropy error computed by:
\begin{equation}
\mathcal{L}_1=-\sum\nolimits_{d = 1}^m {\mathbf{Y}_d^l \ln \mathbf{Z}_d^l }
\end{equation}

Then, the low-layer GCN learns node representations from the node-node-label graph. Similarly, in the first layer each convolution node aggregates information from both the neighborhood common nodes and the associated attributed label nodes (\textit{e.g.}, take node V2 in Fig.~\ref{fig:mlgcn}(a) as an example). Let $\mathbf{E}$ be the $(n+m) \times (n+m)$ adjacency matrix of the input node-node-label graph, the $d_o$-dimensional node embeddings output by the first GCN layer are computed as:
\begin{equation}
\mathbf{N}^{(1)}=\rho (\tilde{\mathbf{E}}^* \mathbf{X}^* \mathbf{W}^v_0)  
\end{equation}
where $\mathbf{W}^v_0 \in \mathbb{R}^{d_i \times m}$ is a weight matrix for the first node-node-label GCN layer, $\mathbf{X}^*=[\mathbf{X};\mathbf{Y}]^T$ and $\tilde{\mathbf{E}}^*$ is a truncated normalized symmetric adjacency matrix obtained by:
\begin{equation}
\mathbf{E}^*=\mathbf{E} + \mathbf{I}_{m};  \mathbf{E}=\mathbf{D}_{e}^{-\frac{1}{2}} \mathbf{E}^{*} \mathbf{D}_{e}^{ - \frac{1}{2}}; \tilde{\mathbf{E}}^*=\tilde{\mathbf{E}}[:n]
\end{equation}
where $\mathbf{D}_e$ is the degree matrix with $\mathbf{D}_{e,ii}=\sum_j \mathbf{E}^*_{ij}$. As in the  label-label-node graph, we can also incorporate $k$-hop neighborhood information by stacking multiple GCN layers:
\begin{equation}
\mathbf{N}^{(k)}=\rho (\tilde{\mathbf{A}} \mathbf{N}^{(k-1)} \mathbf{W}^v_k)
\end{equation}
where $\tilde{\mathbf{A}}=\mathbf{D}_{a}^{-\frac{1}{2}} (\mathbf{A}+\mathbf{I}_m) \mathbf{D}_{a}^{ - \frac{1}{2}}$ and $\mathbf{D}_{a,ii}=\sum_j (\mathbf{A}+\mathbf{I}_n)_{ij}$. The node embeddings output by the last layer have size $m$ and are passed through a sigmoid transformation to perform supervised multi-label classification with the collective cross-entropy loss (\textit{e.g.}, the two-layer GCN are used in this paper) over all labeled nodes computed by:
\begin{equation}
\mathbf{O}^v=\tilde{\mathbf{A}} ReLU (\tilde{\mathbf{E}}^* \mathbf{X}^* \mathbf{W}^v_0)\mathbf{W}^v_1
\end{equation}
\begin{equation}
\mathcal{L}_2=-\sum\nolimits_{i \in \mathbf{y}} \mathcal{L}_2^*
\end{equation}
where $\mathbf{W}^v_0 \in \mathbb{R}^{d_i \times d_h}$ and $\mathbf{W}^v_1 \in \mathbb{R}^{d_h \times m}$ are the weight matrices for the first and second node-node-label GCN layers, respectively, $\mathbf{y}$ is the set of node indices that have labels. Let $\mathbf{Y}^{v}$ be the one-hot label indicator matrix of all common nodes, then $\mathcal{L}_2^{*}$ is calculated:
\begin{equation}
\begin{aligned}
 \mathcal{L}_2^*  &= {\rm \mathbf{Y}}_{i}^{v} \log (\sigma(\mathbf{O}_i^v )) + (1 - {\rm \mathbf{Y}}_{i}^{v} )\log (1 - \sigma(\mathbf{O}_i^v )) \\
                  &= {\rm \mathbf{Y}}_{i}^{v} \log \left( {\frac{1}{{1 + \exp ( - \mathbf{O}_i^v )}}} \right) \\
                  &+\left( {1 - {\rm \mathbf{Y}}_{i}^{v} } \right)\log \left( {\frac{{\exp ( - \mathbf{O}_i^v )}}{{1 + \exp ( - \mathbf{O}_i^v )}}} \right) \\
                  &=  - {\rm \mathbf{Y}}_{ i}^{ v} \log \left( {1 + \exp ( - \mathbf{O}^v_i )} \right) \\
                  \null\hfill &- \left( {1 - {\rm \mathbf{Y}}_{i}^{ v} } \right)\log \left( {\mathbf{O}_i^v  + \log \left( {1 + \exp ( - \mathbf{O}_i^v )} \right)} \right) \\
                  &=  - \left( {1 - {\rm \mathbf{Y}}_{i}^{v} } \right)\mathbf{O}_i^v  - \log (1 + \exp ( - \mathbf{O}_i^v )).
\end{aligned}
\end{equation}

\begin{algorithm}[t]
\SetAlgoLined
\SetKwInOut{Input}{Input}\SetKwInOut{Output}{Output}
\Input{A multi-label graph $G=(\mathbf{v,e,l,A,B,X})$}
\Output{The node representations $\mathbf{O}^n=\{h_i\}_{i=1,\cdots,n}$}
\textbf{Initialization}: $i=0$, the training epoch $I$, the information updating frequencies $M$ and $N$
\BlankLine
\While{$i \le I$}{
    Feed the label-label-node graph to train label representations\;
    Feed the node-node-label graph to train node representations\;
    \If{$i \% M = 0$}{
        Update the feature matrix by Eq. (15)\;
        }
    \If{$i \% N = 0$}{
        Update the feature matrix by Eq. (16)\;
        }
    Optimize $\mathcal{L}_1$ and $\mathcal{L}_2$ by the collective classification objective of Eq. (17);\\
    $i$ = $i$ +1.
}
\caption{Training ML-GCNs\label{EMDalg}}
\end{algorithm}

The above two aspects of representation learning for labels and nodes are trained together and impact one another by sharing the common classification labeling space of $\mathbf{l}$ from the target graph $G$, and in the meantime a subset of input features, \textit{i.e.}, through the attributed label nodes in the low-level node-node graph and the attributed common nodes in the high-level label-label graph. Let the total training epoch for ML-GCN be $I$, after $N$-epoch training of common node representations, the input feature matrix for the label-label-node graph will be updated:
\begin{equation}
\mathbf{X}_{new}=\rho(\mathbf{O}^v \mathbf{W}^v);  \mathbf{Y}^*=[\mathbf{Y};\mathbf{X}_{new}]^T,
\end{equation}
and in the meantime, after $M$-epoch training of label representations, the input feature matrix for the node-node-label graph will be updated:
\begin{equation}
\mathbf{Y}_{new}=\rho(\mathbf{O}^l \mathbf{W}^l);  \mathbf{X}^*=[\mathbf{X};\mathbf{Y}_{new}]^T
\end{equation}
where $\mathbf{W}^v \in \mathbb{R}^{m \times d_i}$ and $\mathbf{W}^l \in \mathbb{R}^{m \times d_i}$ are weight matrices. The collective training procedure for the ML-GCN model has been summarized in Algorithm~\ref{EMDalg}.
\subsection{Algorithm Optimization and Complexity Analysis}
As can be seen from Algorithm 1, the node representations and label representations are not learned independently, but depend on each other through shared embedding features learned from two reciprocally enhanced GCNs. In addition, the two level GCNs conduct two supervised classification tasks within the same labelling space: the top label-label-node GCN is doing a single-label classification and the bottom node-node-label GCN is doing a multi-label node classification. Finally, The global learning objective is to minimize the following collective classification loss:
\begin{equation}
\mathcal{L}=\mathcal{L}_1 + \mathcal{L}_2
\end{equation}
In this paper, all weight parameters are optimized using gradient descent as in \cite{14} and  \cite{25}.

The training of ML-GCN is efficient in terms of the computational complexity. In this paper, we adopt a two-layer GCN and one-layer GCN for learning the node and feature representations, respectively. Since multiplication of the adjacency matrix (e.g., $\mathbf{A}$ for the node-node graph and $\mathbf{F}$ for the label-label graph) and feature matrix (e.g., $\mathbf{Y}^{*}$ and $\mathbf{X}^{*}$ in Eqs. (6) and (12), respectively) can be implemented as a product of a sparse matrix with a dense matrix, the algorithm complexity of ML-GCN can be represented as $O((Ed_id_hm+nd_i)+(Ld_im+md_i))$, where $n$ and $m$ are the number of nodes and labels, $E$ and $L$ are the number of edges in node-node-label graph and label-label-node graph, respectively. $d_i$ is the dimension (for both nodes and labels) of input feature vectors. $d_h$ is dimension of the hidden feature vectors produced in the first node-node-label GCN layer of all common nodes. In addition, because for most networks  $m$, $L$ and $n$ are generally far more less than $E$ (\textit{e.g.}, as we will see in section V, for the Filckr dataset, $E$ is 4,332,620, compared with $m$, $L$ and $n$ are merely 194, 3,716 and 8,052, respectively), therefore the complexity of ML-GCN is approximately equivalent to $O(Ed_id_hm)$, which is the same as GCN. Meanwhile, since Eqs. (15) and (16) are not computed in each epoch (\textit{e.g.}, every 50 epoch), the complexity for our model is still $O(Ed_id_hm)$, the same theoretical asymptotic complexity as the GCN.

\section{Experiments \& Results}
In this section, we compare the proposed approach against a set of strong baselines on three real-world datasets by conducting supervised node classification.

\subsection{Benchmark Datasets}
We collect three multi-label networks \cite{12,21}, BlogCatalog, Flickr, and YouTube, as the benchmark. They are described as follows.

\textbf{BlogCatalog} is a network of social relationships among 10,312 blogger authors (nodes), where the node labels represent bloggers' interests such as \textit{Education}, \textit{Food} and \textit{Health}. There are 39 unique labels in total and each node may be associated with one or multiple labels. It is easy to find that users' labels of interest often interact and correlate with each other to enhance the affinities between blogger authors. For example, \textit{food} is highly related with \textit{Health} in real life, where two users have both labels \textit{food} and \textit{life} should be much closer compared with those whom only share either label \textit{food} or label \textit{life}. There are 615 co-occurrence relationships (\textit{e.g.}, correlations) among all 39 labels in this dataset.

\textbf{Flickr} is a photo-sharing network between users, where node labels represent user interests, such as \textit{Landscapes} and \textit{Travel}. There are 8,052 users and 4,332,620 interactions (\textit{e.g.}, edges) among them. Each user could have one or multiple labels of interest from the same labeling space of 194 labels in total. 

\textbf{YouTube} is a social network formed by video-sharing behaviors, where labels represent the interest groups of users who enjoy common video genres such as \textit{anime} and \textit{wrestling}. Table~\ref{tab:data} summaries their detailed statistic information. There are 22,693 users and 192,722 links between them. Each pair of linked users may share multiple identical labels out of the total 47 labels. The number of correlations between these labels is 1,079.

The detailed statistic information of the above three multi-label networks are summarized in Table 1.

\subsection{Comparative Methods}
We compare the performance of the proposed method with the following state-of-the-art methods for multi-label node classification:
\begin{itemize}
    \item \textbf{DeepWalk} \cite{21} is a shallow network embedding model that only preserves the topology structures. It captures the node neighborhood relations based on random walks and then derives node representations based on SkipGram model.
    \item \textbf{LINE} \cite{23} is also a structure preserving method. It optimizes a carefully designed objective function
    that preserves both the local and global network structures, compared with the DeepWalk that encodes only the local structures.
    \item \textbf{Node2vec} \cite{26} adopts a more flexible neighborhood sampling process than DeepWalk to capture the node relations. The biased random walk of Node2vec can capture second-order and high-order node proximity for representation learning.
    \item \textbf{GENE} \cite{12} is a network embedding method that simultaneously preserves the topology structures and label information. Different from the proposed approach in this paper, GENE simply models labels as plain attributes to enhance structure-based representation learning process, whereas our model considers multi-label correlation and network structure for representation learning. 
    \item \textbf{GCN} \cite{14} is a state-of-the-art method that can naturally learn node relations from network structures and features, where each node forms its representation by adopting a spectral-based convolutional filter to recursively aggregate features from all its neighborhood nodes.
    \item \textbf{Text GCN} \cite{25} is built on GCN that aims to embed heterogeneous information network. In this paper, we construct  heterogeneous node-label graph, where common nodes and label nodes are directly connected by their labeling relationships.
    \item \textbf{ML-GCN$_{node}$} is a variant of the proposed ML-GCN model that removes attributes of common nodes from the label-label-node graph. Therefore, the community information is not preserved in this method.

    \item \textbf{ML-GCN$_{1n}$} is a variant of the proposed ML-GCN model. The only difference with ML-GCN is that ML-GCN$_{1n}$ takes only one convolutional layer while learning from the node-node-label  graph.
    \item \textbf{ML-GCN$_{2l}$} is a variant of the proposed ML-GCN model, which adopts a two consecutive convolutional layers to learn from the label-label-node graph, compared with one layer in GCN.
    \item \textbf{ML-GCN} is our proposed multi-label learning approach in this paper. It considers a two-layer graph structure--a high-level label-label-node graph which allows the preservation of label correlations and meanwhile a low-level node-node-label graph that enables the label correlation-enhanced node representation learning.
\end{itemize}

\begin{table}
\renewcommand{\arraystretch}{1.30}
\centering
\caption{Dataset characteristics.}
\label{tab:data}
    \begin{tabular}{c|c|c|c}
    \hline
    Items & BlogCatalog & Flickr & YouTube \\
    \hline
    \# Nodes & $10,312$  & $8,052$ & $22,693$ \\
    \hline
    \# Edges & $333,983$ & $4,332,620$ & $192,722$ \\
    \hline
    \# Labels & $39$ & $194$ & $47$ \\
    \hline
    \# Co-occur. & $615$ & $3,716$ & $1,079$ \\
    \hline
    \end{tabular}
\end{table}

\

\begin{table*}[ht]
\renewcommand{\arraystretch}{1.30}
\centering
\caption{Multi-label classification performance comparison. The $1^{\text{st}}$, $2^{\text{nd}}$, and $3^{\text{rd}}$ best results are bold-faced, italic-formatted and underscored respectively.}
\label{tab:results}
\begin{tabular}{c|c|c|c|c|c|c|c}
\toprule
\multicolumn{2}{c}{Metrics} & \multicolumn{3}{c}{Micro-F1 (\%)} & \multicolumn{3}{c}{Macro-F1 (\%)} \\
\hline
\multicolumn{2}{c|}{Datasets} & BlogCatalog & Flickr & YouTube & BlogCatalog & Flickr & YouTube\\
\hline
\multirow{10}{4em}{Methods} & DeepWalk & {$29.19${\tiny $\pm0.28$}} & {$25.75${\tiny $\pm0.13$}} &   {$26.19${\tiny $\pm0.18$}} & {$19.22${\tiny $\pm0.63$}} & {$12.31${\tiny$\pm0.17$}} &  {$10.03${\tiny$\pm0.23$}} \\

& LINE & {$30.79${\tiny $\pm0.84$}} & {$30.13${\tiny $\pm0.39$}} &   {$27.68${\tiny $\pm0.11$}} & {$17.89${\tiny $\pm1.26$}} & {$16.24${\tiny$\pm0.53$}} &  {$10.90${\tiny$\pm0.30$}} \\

& Node2vec & {$33.35${\tiny $\pm0.69$}} & {$34.51${\tiny $\pm0.44$}} &   {$26.75${\tiny $\pm0.25$}} & {$21.38${\tiny $\pm1.76$}} & {$19.72${\tiny$\pm0.32$}} &  {$10.03${\tiny$\pm0.29$}} \\

& GENE & {$28.77${\tiny $\pm0.02$}} & {$29.44${\tiny $\pm0.18$}} &   {$26.77${\tiny $\pm0.22$}} & {$16.00${\tiny $\pm1.06$}} & {$14.35${\tiny$\pm0.82$}} &  {$10.48${\tiny$\pm0.72$}} \\

& GCN & {$43.57${\tiny $\pm0.11$}} & \underline{{$40.36${\tiny $\pm0.04$}}} &   $\textit{44.44}${\tiny $\pm\textit{0.02}$} & {$36.55${\tiny $\pm0.22$}} & \underline{{$24.77${\tiny$\pm0.09$}}} &  {$33.54${\tiny$\pm0.08$}} \\

& Text GCN & {$40.31${\tiny $\pm0.49$}} & $\textit{41.82}${\tiny $\pm\textit{0.33}$} &   {$39.82${\tiny $\pm0.53$}} & {$32.03${\tiny $\pm0.49$}} & {$22.98${\tiny$\pm0.40$}} &  {$\textit{39.07}${\tiny$\pm\textit{0.11}$}} \\

& ML-GCN$_{node}$ & \underline{{$43.72${\tiny $\pm0.31$}}} & {$39.99${\tiny $\pm0.07$}} & \underline{$44.14${\tiny $\pm0.05$}} & \underline{{$38.39${\tiny $\pm0.38$}}} & {$22.70${\tiny$\pm0.03$}} &  \underline{{$33.96${\tiny$\pm0.18$}}} \\

& ML-GCN$_{1n}$ & {$32.96${\tiny $\pm0.01$}} & {$31.83${\tiny $\pm0.03$}} &   {$31.82${\tiny $\pm0.07$}} & {$17.92${\tiny $\pm0.02$}} & {$16.89${\tiny$\pm0.21$}} &  {$27.48${\tiny$\pm0.33$}} \\

& ML-GCN$_{2l}$ & {$\textit{43.86}${\tiny $\pm\textit{0.07}$}} & {$38.41${\tiny $\pm0.23$}} &   {$42.33${\tiny $\pm0.03$}} & {$\textit{38.63}${\tiny $\pm\textit{0.24}$}} & {$\textit{26.28}${\tiny$\pm\textit{0.30}$}} &  {$32.62${\tiny$\pm0.11$}} \\

& ML-GCN & {$\mathbf{45.17}${\tiny $\pm\mathbf{0.20}$}} & {$\mathbf{43.74}${\tiny $\pm\mathbf{0.20}$}} &   {$\mathbf{45.71}${\tiny $\pm\mathbf{0.02}$}} & {$\mathbf{42.53}${\tiny $\pm\mathbf{0.72}$}} & {$\mathbf{30.71}${\tiny$\pm\mathbf{0.12}$}} &  {$\mathbf{42.77}${\tiny$\pm\mathbf{0.31}$}} \\
\bottomrule
\end{tabular}
\end{table*}

The above baselines can be roughly separated into three categories based on the types of information (\textit{e.g.}, network structures and labels) and how it is incorporated in the graph embedding models. The first class belongs to methods that only preserve graph structures, including DeepWalk, Node2vec, LINE, GCN (\textit{e.g.}, we use the structure-based identity matrix as the original features of all node). The second class includes GENE and Text GCN that preserve both the graph structures and label information, where the labels are modeled as plain attribute information to enhance structure-based representation learning. The proposed method MG-GCN and its variants (ML-GCN$_{node}$, ML-GCN$_{1n}$, and ML-GCN$_{2l}$) represent the thrid class, which not only preserve structural and label information, but also the correlations between labels. 

It is worth noting that we designed three variants of MG-GCN (including ML-GCN$_{node}$, ML-GCN$_{1n}$, and ML-GCN$_{2l}$) to validate its performance under different settings. This allows us to fully observe ML-GCN's performance and conclude which part is playing major roles for multiple-label GCN learning. 

\subsection{Experiment Setup}

There are many hyper-parameters involved. Some are empirically set \cite{25} while others are selected through sensitivity experiments. For ML-GCN, we use two-layer and one-layer GCNs to learn from the node-node-label graph and the label-label-node graph respectively. We test hidden embedding size, $d_h$, between 50 to 500, training ratios, $\alpha$, of supervised labeled instances between 0.025 and 0.2, and updating frequencies $N$ and $M$ from 10 to 100, respectively. We also compare the performance of ML-GCN through differing numbers of GCN convolutional layers (e.g., ML-GCN$_{1n}$ and ML-GCN$_{2l}$). For comparison, we set the learning rate $\eta$ for gradient decent as 0.02, training epoch as 300, dropout probability as 0.5, $L_2$ norm regularization weight decay as 0, and the default values of $d_h$, $\alpha$, $N$ and $M$ as 400, 0.2, 50 and 50, respectively. After selecting the labeled training instances, the rest is split into two parts: 10\% as validation set and 90\% for testing set.

It is necessary to mention that all baselines are set to conduct the multi-label node classification (e.g., each node can belong to multiple labels) within the same environmental settings. As metrics used in \cite{21} and \cite{23}, we adopt Micro-F1 and Macro-F1 to evaluate the node classification performance, which are defined as follows:
\begin{equation}
{\rm Micro-F1} = \frac{{\sum\nolimits_{i = 1}^\mathbf{l} {2TP^i } }}{{\sum\nolimits_{i = 1}^\mathbf{l} {(2TP^i  + FP^i  + FN^i )} }}
\end{equation}
\begin{equation}
{\rm Macro-F1} = \frac{1}{\mathbf{l}}\sum\nolimits_{i = 1}^\mathbf{l} {\frac{{2TP^i }}{{\left( {2TP^i  + FP^i  + FN^i } \right)}}}
\end{equation}
where $\mathbf{l}$ is the set of labels from the target graph \textit{G}. $TP^i$, $FN^i$ and $FP^i$ denote the number of true positives, false negatives and false positives w.r.t the $i$th label category, respectively. All experiments are repeated 10 times with the average results and their standard deviations reported.

\begin{figure*}[t]
\centering
\begin{minipage}[b]{0.33\linewidth}
\centering 
\includegraphics[width=0.893\textwidth]{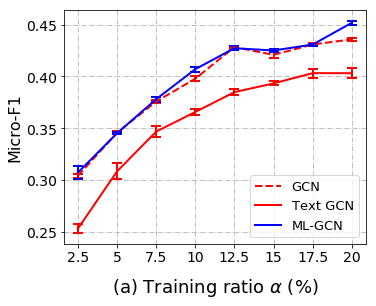}
\end{minipage}%
\hspace{10mm}
\begin{minipage}[b]{0.33\linewidth}
\centering 
\includegraphics[width=0.893\textwidth]{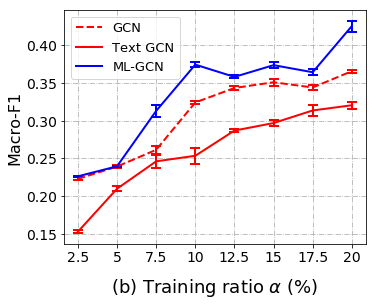}
\end{minipage}%
\caption{Algorithm performance comparisons with respect to different percentage of training sample ratios (the $x-$axis denotes the ratio of training samples comparing to the whole network).}
\end{figure*} 

\subsection{Experimental Results}
Table~\ref{tab:results} presents the comparative results of all methods with respect to the multi-label classification performance within the same environment settings, where the top three best results have been highlighted. From the table, we have the following four main observations.
\begin{itemize}
    \item Among all methods that encode only the graph topology structures, the shallow neural networks-based methods (e.g., DeepWalk, LINE and Node2vec) perform poorly with a wide gap compared with deep model GCN over all three datasets, \textit{i.e.}, on BlogCatalog network, the classification performance of GCN improved 30.6\% and 70.9\% over Node2vec w.r.t Micro-F1 and Macro-F1, respectively. This is because shallow models have limitations in learning complex relational patterns among nodes \cite{zhang2018network}. For example, although Node2vec relies on a carefully designed random walk process to capture the node neighborhood relationships, it cannot differentiate the affinities between a node and others within the same walk sequence. In comparison, GCN uses a more efficient way to constrain the neighborhood relations between nodes, where each node only interact with its neighbors in each convolution layer. Such a learning paradigm is more accurate to maintain the actual node relevance reflected by the edge links without introducing noise neighborhood relationships as in Node2vec.
    \item In terms of the performance of methods (GENE and Text GCN) that have incorporated the labels to enhance the structures modeling. GENE is built on DeepWalk to additionally preserve the label information. We can observe that GENE performs slightly better than DeepWalk on both Flickr and YouTube datasets. The reason is that each label is considered as the higher-level summation of a group of similar nodes, thus can be used to supervise and distinguish the neighborhood affinities between nodes within the same random walk sequence. Nevertheless, LINE and Node2vec can perform better than GENE in most cases over three datasets, \textit{i.e.}, the Micro-F1 performance of LINE and Node2vec on BlogCatalog increased 2.0\% and 4.6\%, respectively. The reason is probably that they have adopted more efficient random walk process to capture node neighborhood relationships. In addition, as we can see from Table~\ref{tab:results}, the label-preserved model, Text GCN, has no advantages compared with the basic GCN model. The reason is probably that the labels are considered as attributes that have not been leveraged in a meaningful manner. In other words, only the labeled nodes have the attribute of labels in the supervised node representation leaning and training, the scattered labels could have become the noisy information to confuse the neighborhood relationships modeling between nodes.

\begin{figure*}[t]
\centering
\begin{minipage}[b]{0.25\linewidth}
\centering 
\includegraphics[width=1\textwidth]{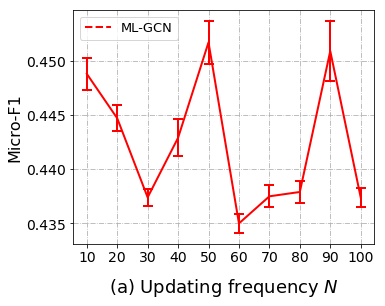}
\end{minipage}%
\begin{minipage}[b]{0.25\linewidth}
\centering 
\includegraphics[width=1\textwidth]{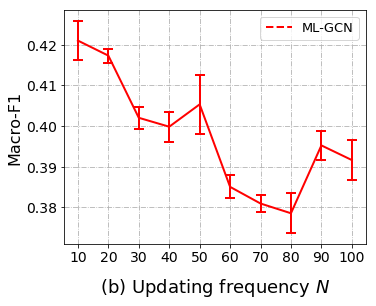}
\end{minipage}%
\begin{minipage}[b]{0.25\linewidth}
\centering 
\includegraphics[width=1\textwidth]{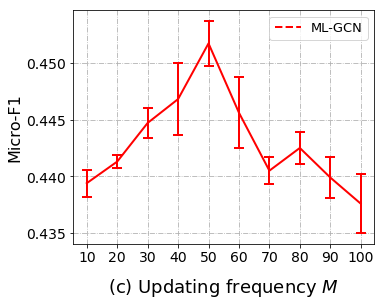}
\end{minipage}%
\begin{minipage}[b]{0.25\linewidth}
\centering 
\includegraphics[width=1\textwidth]{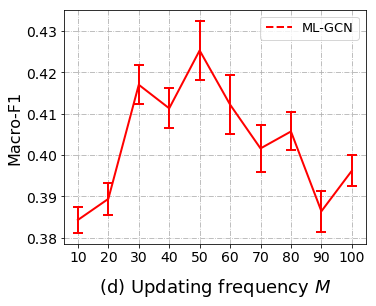}
\end{minipage}%
\caption{Impact of the information updating frequencies.}
\end{figure*}

\begin{figure}[t]
\centering
\begin{minipage}[b]{0.5\linewidth}
\centering 
\includegraphics[width=1\textwidth]{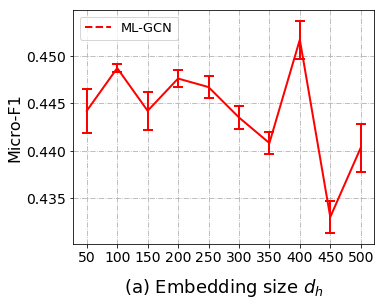}
\end{minipage}%
\begin{minipage}[b]{0.5\linewidth}
\centering 
\includegraphics[width=1\textwidth]{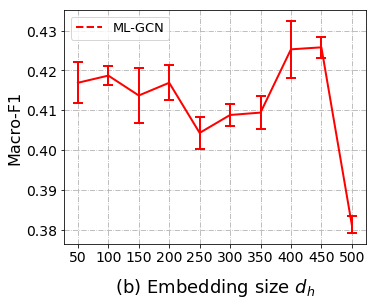}
\end{minipage}%
\caption{Impact of the hidden node embedding size.}
\end{figure}    
    \item For deep models that both preserved graph structures, labels and label correlations, we can observe that ML-GCN is consistently superior to the Text GCN in leaning multi-label graphs over all three datasets. The reasons are mainly in three-fold. First, ML-GCN allows immediate and efficient label correlation modeling without depending on common nodes, \textit{i.e.}, labels directly interact with each other over the label-label graph in the proposed model. Second, it is common to use different numbers of convolutional layer (e.g., based on the results observed by comparing ML-GCN and ML-GCN$_{2l}$, where the former performs far more better) to learn from the label and node graphs respectively, since the node-node graph is much more complicated than the label-label graph that involves simple label interaction patterns. To obtain a model that best fits the given label and node graphs of different scales, one can easily change the number of layers used by the two-layer graph modelings in ML-GCN independently. But this is hard  for Text GCN to coordinate the layer settings that are most suitable to model node relations and label relations simultaneously in the node-label graph. Finally, in the structure design of ML-GCN, each label has preserved the community information by taking all related common nodes as attributes, which make the node relations modeling and the label correlations modeling more dependent on each other to optimize the global network representation learning, \textit{i.e.}, the node representations of one label could refine the node representation learning of another correlated label, as we will demonstrate in the case study later.
  \item In terms of the performance of different variants for the proposed ML-GCN, we can condlude that Ml-GCN is superior to ML-GCN$_{node}$, ML-GCN$_{1n}$ and ML-GCN$_{2l}$, where the possible reasons are given as follows: 1) The comparison between ML-GCN and ML-GCN$_{node}$ demonstrates that taking the common nodes as attributes of the label-label network is beneficial, where labels could learn more enriched representations (with encoding label correlations and the node communities) to refine the neighborhood feature aggregation for node representation learning in the low-level node-node graph; 2) ML-GCN performs better than ML-GCN$_{2l}$, which illustrates a single-layer GCN is appropriate to model the label correlations, since compared with the node-node interactions, the label-label interactions are generally simple and explicit; 3) ML-GCN$_{2l}$ is poorly performed than ML-GCN, which demonstrates that exploring the high-order neighborhood relationships between nodes is important.
\end{itemize}

\subsection{Parameter Sensitivity}
We designed extensive experiments to test the sensitivities of various parameters between a wider range of values, such as the training ratio $\alpha$ of labeled nodes, the feature updating frequencies $N$ and $M$, and the embedding size of the first hidden convolution layer while modeling the node-node-label graph. Fig. 3 shows the impacts of different portions of labeled training instances. In general, for all test models, we can observe that both the Micro-F1 (\textit{e.g.}, Fig. 3(a)) and Macro-F1 (\textit{e.g.}, Fig. 3(b)) performances increase with more labeled training nodes. This is reasonable since all these models adopt a supervised node representation learning and training manner, where the model parameters can be fully trained with larger labeled data \cite{25}. Fig. 4 shows the influence of the input feature-updating frequencies  controlled by $N$ and $M$ (\textit{e.g.}, used in Eqs. (15) and (16)). We can see from Fig. 4(a) where the performance changes with $N$ but no clear patterns can be observed in Micro-F1, while Fig. 4(b) shows an deceasing trend with larger values of $N$ in the Macro-F1 scores. In comparison, from Fig. 4(c) and Fig. 4(d) the accuracy first increases then decreases with larger values of $M$ w.r.t. both Micro-F1 and Macro-F1 scores. We also test the impact of node embedding size generated by the first convolutional layer with the trend shown in Fig. 5(a). The accuracy fluctuates before peaking at 400 and 450 w.r.t. Micro-F1 and Macro-F1 results, followed by a sharp decline.

\subsection{Case Study}
To illustrate how label correlations affect the multi-label graph learning performance, we present the classification results through four related label categories shown in Fig. 6. Fig. 7 presents their correlation matrix, where deeper colors imply higher correlation between two corresponding labels, \textit{i.e.}, L1 and L2 are highly correlated. We can see in Fig. 6 that ML-GCN and GCN perform similarly with respect to the node classification of category L1. However, the accuracy of MC-GCN improved over GCN with respect to categories L2, L3 and L4. It is interesting to note how much these categories improved (\textit{e.g.}, L2 $>$ L3 $>$ L4) is related to how frequently they correlate with label L1 respectively. This phenomenon might be caused by the fact that the L1 has an impact on its correlated labels during training. This also verifies that label interaction is critical for multi-label graph learning, and our proposed ML-GCN model can effectively capture and utilize this property.

\section{Conclusions} 
In this paper, we formulated a new multi-label network representation learning problem, where each node of the network may have multiple labels. To simultaneously explore label-label correlation and the network topology, we proposed a multi-label graph convolution network (ML-GCN) to build two siamese GCNs, a node-node-label graph and a label-label-node graph from the multi-label network, and carry out learning of node representation and label representation from the two GCNs simultaneously. Because the two GCNs are unified to achieve one optimization goal, the learning of node representation and label representation can mutually benefit each other for maximum performance gain. Experiments on three real-word datasets verified the effectiveness of ML-GCN in combining labels, label correlations, and graph structures for enhanced node representation learning and classification. 

\begin{figure}[t]
\centering
\begin{minipage}[b]{0.59\linewidth}
\centering 
\includegraphics[width=1\textwidth]{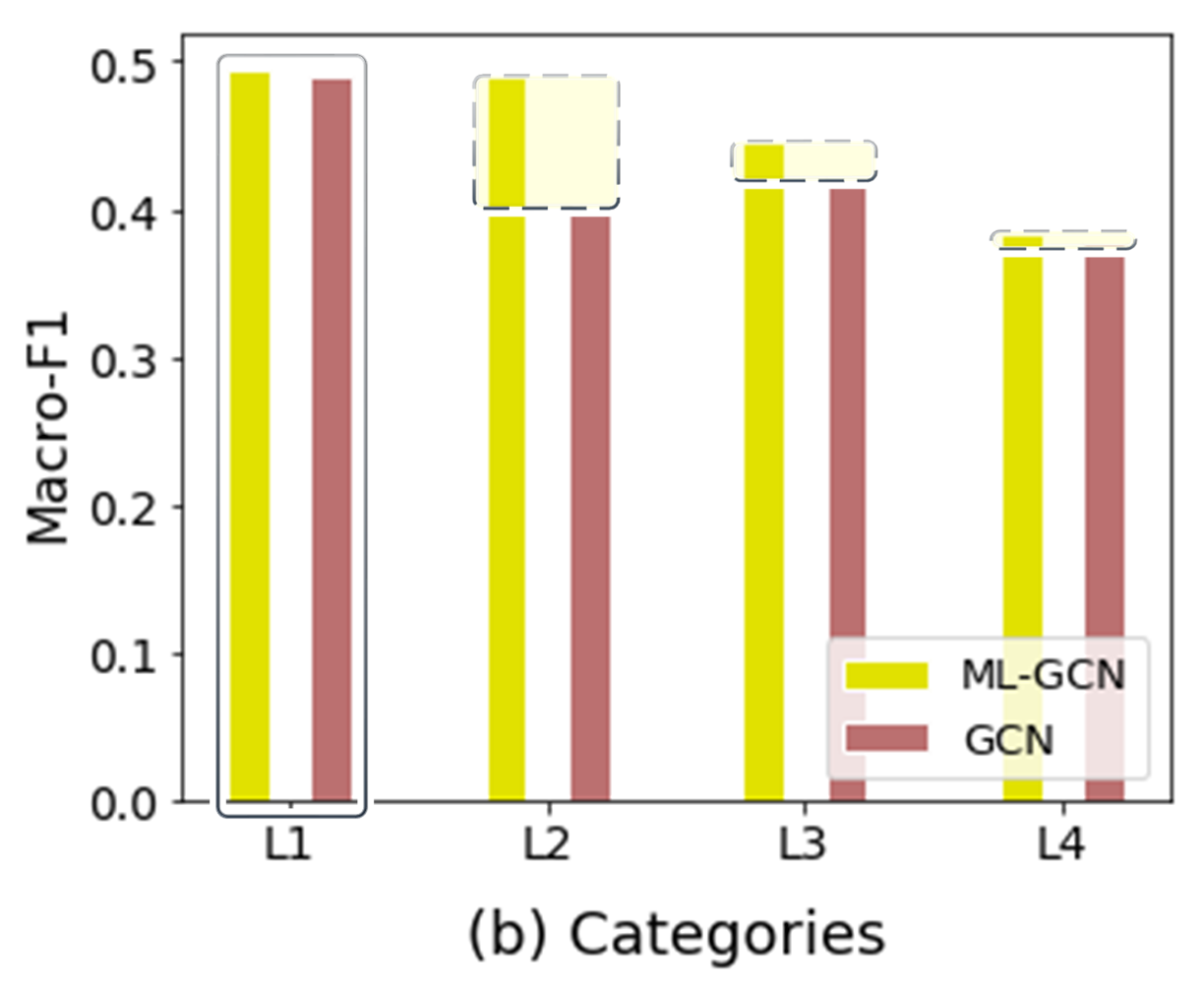}
\end{minipage}%
\caption{Classification performance (Macro-F1) with respect to different label categories.}
\label{fig:5}
\end{figure}

\begin{figure}[t]
\centering
\begin{minipage}[b]{0.59\linewidth}
\centering 
\includegraphics[width=1\textwidth]{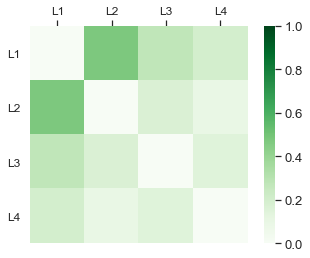}
\end{minipage}%
\caption{Pair-wise label correlation matrix. A higher gray intensity value (excluding main diagonal values) indicates a stronger correlation between two labels.}
\label{fig:5}
\end{figure}




\bibliographystyle{IEEEtran}
\bibliography{references}

\end{document}